\documentclass{article}
\usepackage{microtype}
\usepackage{graphicx}
\usepackage{subcaption}
\usepackage{multirow}
\usepackage{booktabs}
\usepackage{hyperref}

\usepackage[accepted]{icml2025}
\usepackage{amsmath}
\usepackage{amssymb}
\usepackage{mathtools}
\usepackage{amsthm}

\usepackage[capitalize,noabbrev]{cleveref}
\theoremstyle{plain}

\theoremstyle{definition}

\theoremstyle{remark}

\usepackage[textsize=tiny]{todonotes}

\begin{document}

\twocolumn[
\icmltitle{Evaluating Deepfake Detectors in the Wild}
\icmlsetsymbol{equal}{*}

\begin{icmlauthorlist}
\icmlauthor{Viacheslav Pirogov}{equal,sumsub}
\icmlauthor{Maksim Artemev}{equal,sumsub}
\end{icmlauthorlist}

\icmlaffiliation{sumsub}{Sumsub, Berlin, Germany}

\icmlcorrespondingauthor{Viacheslav Pirogov}{slava.pirogov@sumsub.com}

\icmlkeywords{Deepfake Detection, Face Swapping, Model Evaluation}

\vskip 0.3in]

\printAffiliationsAndNotice{\icmlEqualContribution}
\begin{abstract}
  Deepfakes powered by advanced machine learning models present a significant and evolving threat to identity verification and the authenticity of digital media.
  Although numerous detectors have been developed to address this problem, their effectiveness has yet to be tested when applied to real-world data.\\
  In this work we evaluate modern deepfake detectors, introducing a novel testing procedure designed to mimic real-world scenarios for deepfake detection. Using state-of-the-art deepfake generation methods, we create a comprehensive dataset containing more than 500,000 high-quality deepfake images.\\
  Our analysis shows that detecting deepfakes still remains a challenging task. The evaluation shows that in fewer than half of the deepfake detectors tested achieved an AUC score greater than 60\%, with the lowest being 50\%.
  We demonstrate that basic image manipulations, such as JPEG compression or image enhancement, can significantly reduce model performance. 
  All code and data are publicly available at  \url{https://github.com/SumSubstance/Deepfake-Detectors-in-the-Wild}
\end{abstract}

\section{Introduction}
\label{sec:intro}

Due to recent advances in the field of generative computer vision, many technologies once deemed impossible are now real and accessible to the general public.
High-quality and high-fidelity face generation \cite{wu2023high, liu2021blendgan, Gecer_2021}, as well as precise manipulation of facial attributes \cite{mou2023dragondiffusion, kim2022diffface, hou2022feat}, can be achieved by almost anyone using publicly available open-source code and computing tools such as Google Colab \cite{GoogleColab}.
While this democratization of technology has positive aspects, it also means that fake news \cite{zhou2020fake, huang2023faking},  impersonation \cite{li2020measurementdriven}, false authentication, and liveness bypassing \cite{sabaghi2021deep} is easier to perpetrate than ever. 
In the context of this growing concern, deepfakes — highly realistic digital manipulations of human faces — have emerged as a potent tool in the arsenal of misinformation and identity theft.

However, the AI technology behind deepfake generation can also be used to benefit various fields. For instance, AI-generated videos and images can be used to create interactive and engaging learning experiences \cite{dali}.
However, the potential for misuse of this technology, primarily for deepfake generation, has raised significant concerns. 
In this work, we focus on one of the most prominent deepfake-related risks—convincing but false multimedia content—which can be used to spread misinformation, manipulate public opinion, or perpetrate fraud.

At their core, modern deepfake technologies usually leverage a generative model. Typically, this is a Generative Adversarial Network (GAN) \cite{DBLP:GAN} or one belonging to the more powerful family of diffusion models \cite{sohl2015deep, ho2020denoising}. There are two main options for generating a deepfake: the first and more popular option is to create an entirely synthetic image; \cite{rombach2022high, ramesh2022hierarchical, saharia2022photorealistic} the second,  method is to take two photos of real people and swap their faces, which is easier to produce than the first method \cite{10.1145/3394171.3413630, roop}.

The modern state of deepfake generation and detection \cite{passos2023review, yi2023audio} can be likened to a strategic game of attack and defense, where each side possesses equal strength.
Although it appears that the field is evenly matched, this work aims to provide a closer examination of the capabilities of open-source deepfake detectors.
We introduce minor augmentations that are commonly employed in deepfake production to emulate techniques by fraudsters to bypass security measures. This is to evaluate the robustness of various deepfake detectors.

We separate forged images into two categories: the first are synthetic, AI-generated images; the second are deepfakes, where the goal is to swap the faces of people. To highlight the issue that modern deepfake detection techniques are not adequately tested and that their impressive performance on current benchmarks do not reflect practical applications, we present an image-level testing pipeline that is much closer to real-world data and designed for easy adaptation to various scenarios. Our findings reveal that all tested state-of-the-art deepfake detectors, despite their near-perfect quality on existing datasets, demonstrate significant limitations when applied in more diverse real-world environments. Furthermore, we published all the code for the models, experiments, and data creation pipelines used, along with a deepfake dataset comprising over 500,000 images for further exploration, research, and use.

\section{Related Work}
This section is structured to follow the deepfake life cycle: creating the deepfake \ref{sec:deepfake_gen}, then enhancing its quality \ref{sec:deepfake_enh}, followed by detection \ref{sec:detectors}. There is a broad range of research that evaluates deepfake detectors \cite{liu2024evolving, heidari2024deepfake, le2023facial}. However, the core focus of most of these papers is not on evaluation, with only a few focusing specifically on that aspect. 

\subsection{Deepfake Generation}
\label{sec:deepfake_gen}

The most powerful and resource-intensive deepfake generation technique was pioneered and later improved in DeepFaceLab \cite{perov2020deepfacelab}. 
The method allows for fast and precise face swapping by utilizing a dataset containing unpaired face photos of the source and target. The pipeline proposed for deepfake generation consists of a set of carefully crafted segmentation and keypoints estimation models as well as adversarially trained autoencoders. 
The collection of images required to train such model from scratch is commonly referred to as a faceset, an extensive compilation of photographs showcasing various angles and expressions of both the target’s and source’s faces, often encompassing tens of thousands of images.
The comprehensive nature of a faceset allows the model to meticulously process every facial nuance, tailoring it to a specific individual.
However, this method's effectiveness hinges on the availability of a complete faceset, which can be a limiting factor, especially for non-public figures.
In terms of available models, DeepFaceLab stands out as the most widely recognized and utilized open-source software in the realm of faceset-based deepfake generation.

However, the most widely adopted modern approach, particularly favored by individuals involved in fraudulent activities, is a zero-shot deepfake generation. 
It leverages publicly accessible open-source GAN-based models \cite{10.1145/3394171.3413630, nirkin2019fsgan, nirkin2022fsganv2, li2019faceshifter} that are adept at transferring faces from one individual to another without necessitating additional training.
Although the results are not as refined as those produced by the previous method, they are significantly more convincing than those generated by the facial keypoint alignment technique.
This method is exemplified in various "fun-tech" applications \cite{reface, swapface, vidnoz, faceswapper.ai, pixble}, which are popular for their ease of use and accessibility.
Detecting deepfakes generated by this approach is the main focus of this work.

\textbf{SimSwap: An Efficient Framework For High Fidelity Face Swapping} is a state-of-the-art open-source model for high fidelity face swapping presented in \cite{10.1145/3394171.3413630}.
It employs an adversarially trained encoder-decoder architecture augmented by an Identity (ID) Injection Module. 
The ID Injection Module separates the attributes and identity of an image to transfer information from the source image to decoder.
It also utilizes a face recognition network \cite{DBLP:ArcFace} to extract an identity vector, which is then integrated using Adaptive Instance Normalization (AdaIN) \cite{DBLP:AdaIN}.
The decoder reconstructs the image, with identity preservation quantified by calculating the cosine similarity between the identity vector of the source and the generated image.
To prevent overfitting and ensure that the generated images look real, the authors adopt adversarial training methods \cite{DBLP:GAN, DBLP:FUNIT, DBLP:BigGAN, DBLP:WGAN-GP, DBLP:styleGAN} with a PatchGAN discriminator \cite{DBLP:pix2pix}.
In addition, feature matching loss \cite{DBLP:pix2pixHD} is used, which involves aligning discriminator features from different levels to reduce attribute mismatch. 
The model was trained on the VGGFace2 dataset \cite{DBLP:vggface}, resized to 224x224 pixels.
A higher resolution 512x512 variant of the model was trained on VGGFace2-HQ, an enhanced version of the original dataset.

\begin{figure*}[ht]
\vskip 0.2in
  \begin{center}
  \begin{subfigure}[b]{0.3\linewidth}
    \includegraphics[width=\linewidth]{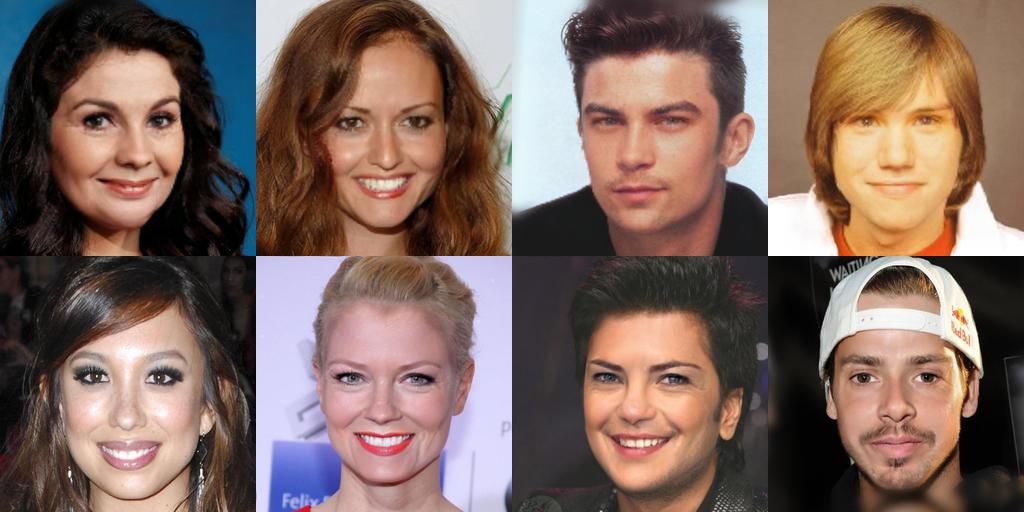}
    \caption{CelebA-HQ}
  \end{subfigure}
  \begin{subfigure}[b]{0.3\linewidth}
    \includegraphics[width=\linewidth]{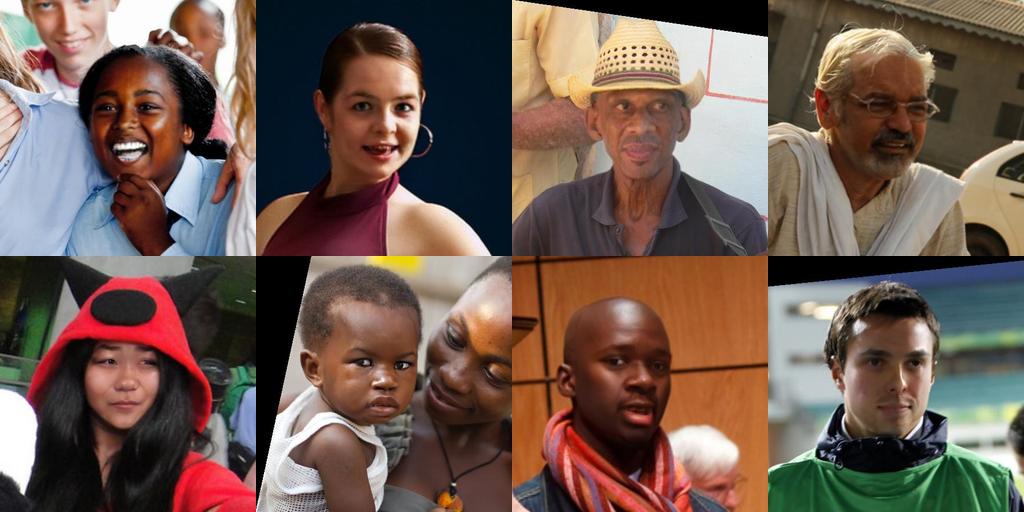}
    \caption{FairFace}
  \end{subfigure}
  \begin{subfigure}[b]{0.3\linewidth}
    \includegraphics[width=\linewidth]{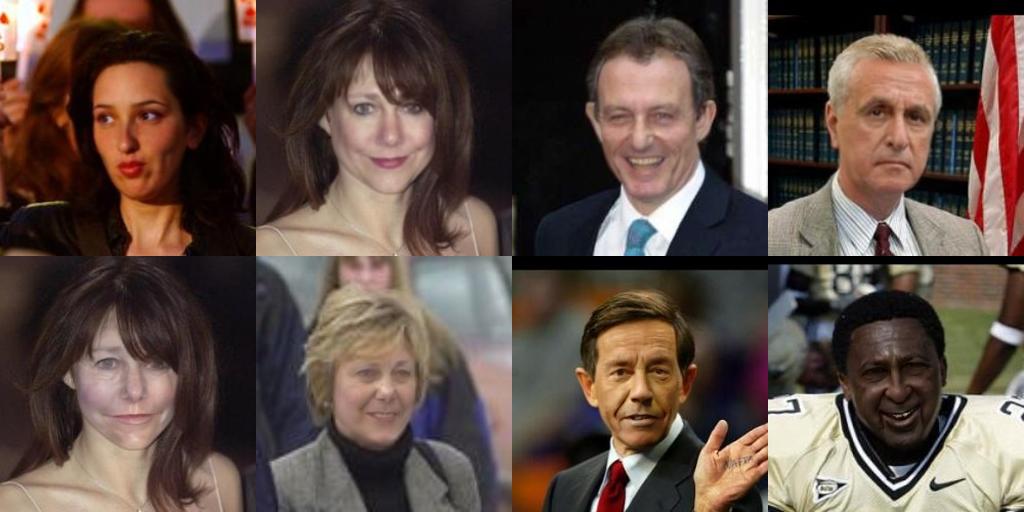}
    \caption{Labeled Faces in Wild}
  \end{subfigure}
    \caption{Samples from our synthetic deepfake dataset. Images in the top row were processed by the Inswapper \cite{InsightFace} (roop) model, while images in the bottom row were generated using SimSwap \cite{10.1145/3394171.3413630}. Samples from Inswapper have slightly lower quality because the model operates at a lower resolution of 128.}
    \label{fig:synth_data}
    \end{center}
\vskip -0.2in
\end{figure*}

\textbf{Inswapper}, developed by InsightFace \cite{InsightFace}, is a notable closed-source model in the field of face swapping.
It is based on a Generative Adversarial Network framework \cite{DBLP:GAN}, however, unlike SimSwap \cite{10.1145/3394171.3413630}, which supports resolutions of 224 and 512, Inswapper operates at a lower resolution of 128.
Despite the lack of publicly available code or detailed model specifications, Inswapper has gained considerable popularity because of its simplicity and ease of use.
It is commonly referred to as "roop" in online forums and repositories, and in 2024, it was integrated into several popular repositories \cite{roop, roop_SD, roop_unleashed, Fooocus_inswapper}. 
Figure \ref{fig:synth_data} presents samples of Inswapper and SimSwap.

\subsection{Deepfake Enhancement}
\label{sec:deepfake_enh}

Image enhancement algorithms such as \textbf{GPEN} \cite{yang2021gan} and \textbf{CodeFormer} \cite{zhou2022towards}, have become popular not only for their intended use of enhancing vintage images, but also for their ability to effectively reduce minor imperfections in the output of other generative models.

GPEN, as presented in the "GAN Prior Embedded Network for Blind Face Restoration in the Wild" paper \cite{10.1145/3594315.3594647}, marks a departure from earlier approaches \cite{Bulat2017SuperFANIF, Li_2018_ECCV, 9025508} which relied primarily on GANs to map degraded faces directly to original images. Instead, GPEN utilizes an external CNN encoder trained to convert degraded images into latent code. The decoder then reconstructs a realistic image from this code, aiming for a one-to-one mapping rather than the conventional one-to-many approach. 
Simultaneously, "Towards Robust Blind Face Restoration with Codebook Lookup Transformer" \cite{zhou2022towards} introduces CodeFormer, an innovative encoder-decoder GAN-like model. It incorporates a discrete codebook integrated into the GAN's latent space, enhancing its capability to restore and refine facial images.

\begin{figure*}[ht]
  \centering
  \begin{subfigure}[b]{0.3\linewidth}
    \includegraphics[width=\linewidth]{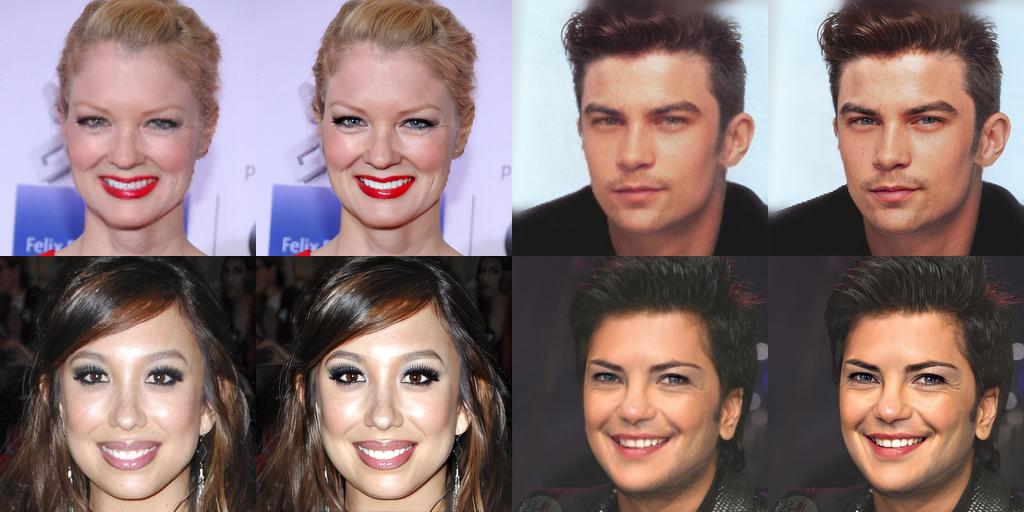}
    \caption{CelebA-HQ}
  \end{subfigure}
  \begin{subfigure}[b]{0.3\linewidth}
    \includegraphics[width=\linewidth]{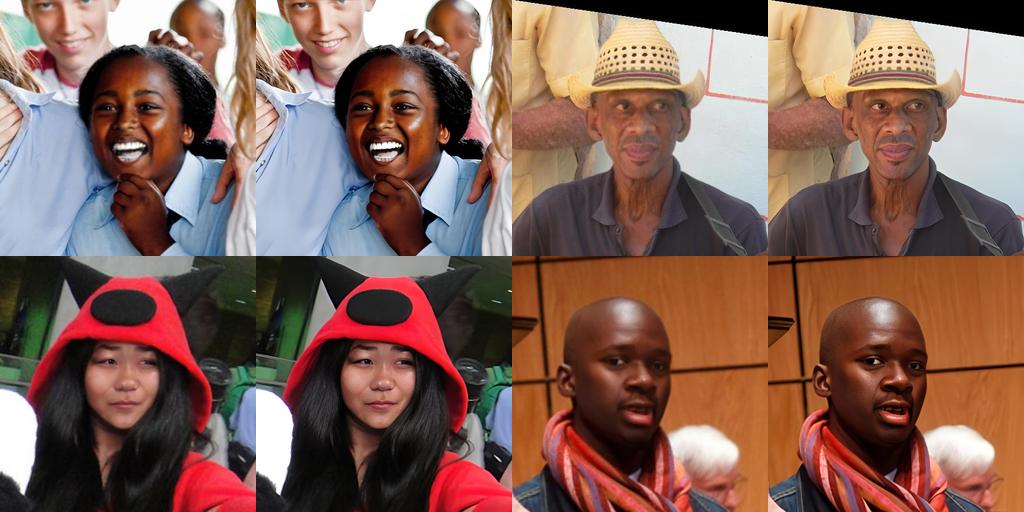}
    \caption{FairFace}
  \end{subfigure}
  \begin{subfigure}[b]{0.3\linewidth}
    \includegraphics[width=\linewidth]{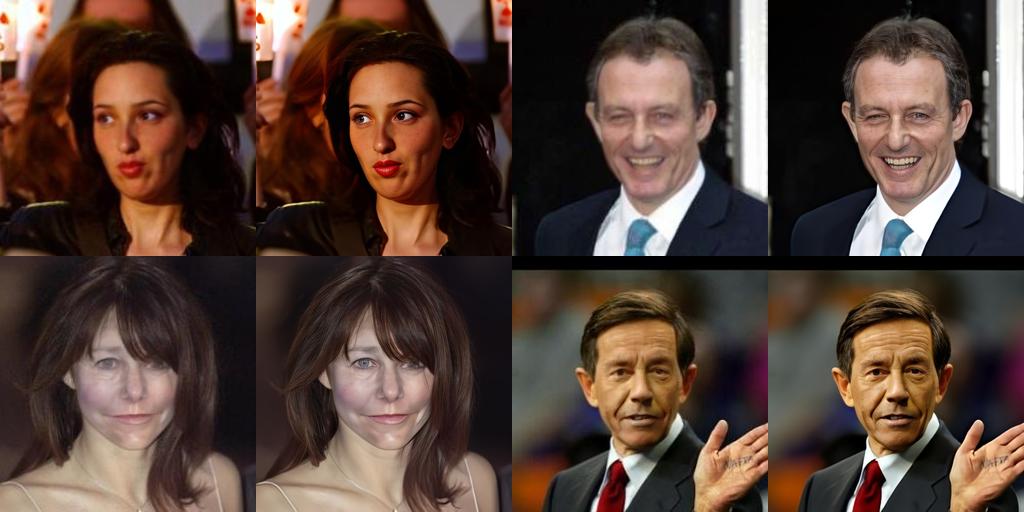}
    \caption{Labeled Faces in Wild}
  \end{subfigure}
    \caption{Samples from our synthetic dataset enhanced using GPEN \cite{10.1145/3594315.3594647}. Originals on the left; enhanced versions on the right. Top: Inswapper outputs; bottom: SimSwap. We demonstrate that enhancers render low-quality generators' outputs visually comparable to real samples.}
    \label{fig:gpen_data}
\end{figure*}

Both GPEN and CodeFormer are adept at correcting minor imperfections such as smoothing skin textures, correcting lighting inconsistencies, and sharpening facial features.
This refinement is crucial in making synthetic faces more lifelike and less artificial.
In particular, these models are community-favored tools for post-processing images from both popular and strong diffusion models \cite{rombach2022high, ramesh2022hierarchical, saharia2022photorealistic}. This is because generating a high-quality face is still a challenging task, which is effectively addressed by the image enhancing capabilities of these models.
As demonstrated in Figure \ref{fig:gpen_data}, it can elevate even low-resolution samples to a visual quality that is indistinguishable from real data.

\subsection{Deepfake Detection}
\label{sec:detectors}

The lack of readily available code and weights in the majority of related work on deepfake detection underscores a very important challenge for researchers and regulators alike.
In this work, we have selected a shortlist of the most popular and diverse open-source approaches to deepfake detection with openly published code and model weights.
The following sections provide brief overviews of these selected approaches.
These methods vary in their focus: Some emphasize innovative model architectures, while others focus on novel training techniques, or creation of synthetic datasets for improved detection.

\noindent\textbf{FaceForensics++} \cite{rossler2019faceforensics++} is a cornerstone in deepfake detection. It offers a dataset of over 1.8 million images from 1000 YouTube videos. The detection model, based on XceptionNet \cite{chollet2017xception} and fine-tuned on this dataset, remains a fundamental benchmark due to its accessibility.

\noindent\textbf{Multi-attentional Deepfake Detection} (MAT) \cite{zhao2021multi} uses a multi-attention network inspired by fine-grained classification and Visual Transformers \cite{dosovitskiy2020image, vaswani2017attention}. It focuses on potential artifact regions within images, combining low-level textural features with high-level semantic features for better detection.

\noindent\textbf{M2TR: Multi-modal Multi-scale Transformers for Deepfake Detection} \cite{wang2022m2tr} integrates multi-modal and multi-scale approaches with a frequency filter \cite{ricker2022towards}, using a 2D Fast Fourier Transform and Cross-Modality Fusion block to merge RGB and frequency domain features, enhancing detection capabilities.

\noindent\textbf{End-to-End Reconstruction-Classification Learning for Face Forgery Detection} (RECCE) \cite{9878441} combines reconstruction learning \cite{wertheimer2021few}, which attunes representations of real images to unknown forgery patterns, with classification learning to mine disparities between real and fake images. This approach uses a multi-scale strategy and modified metric-learning loss for better detection.

\noindent\textbf{Implicit Identity Leakage: The Stumbling Block to Improving Deepfake Detection Generalization} (CADDM) \cite{dong2023implicit} addresses 'Implicit Identity Leakage', where deepfake images blend multiple identities. CADDM swaps specific facial regions and trains a multi-scale detection model to scrutinize feature maps, improving artifact detection.

\noindent\textbf{Detecting Deepfakes with Self-Blended Images} (SBI) \cite{shiohara2022detecting} uses a unique dataset generated through the blending of pseudo-source and target images derived from individual pristine images. This approach replicates common forgery artifacts without relying on any existing deepfake datasets. The EfficientNet-b4 model \cite{tan2019efficientnet}, fine-tuned on this dataset, detects four primary deepfake artifacts: landmark mismatch, blending boundary, color mismatch, and frequency inconsistency.

\section{Datasets}
\label{sec:datasets}

To evaluate the performance of the selected models, we used three public open-source datasets: CelebA-HQ \cite{karras2018progressive}, Labeled Faces in the Wild \cite{LFWTech}, and FairFace \cite{karkkainenfairface}, as described in \ref{sec:face_datasets}. We also created our own deepfake dataset following the procedure outlined in \ref{sec:synth-data}. 
In the subsection \ref{sec:augmentations}, we propose simple augmentation techniques that fraudsters typically use to bypass detectors.

\subsection{Face Datasets}
\label{sec:face_datasets}

The \textbf{CelebA-HQ dataset} \cite{karras2018progressive} specifically designed for training and benchmarking advanced image processing algorithms.
It features a large collection of high-resolution face images of celebrities, who are often a target for deepfake attacks.
Due to the relatively high resolution of images in CelebaHQ, it is often used for training general deepfake generators.

The \textbf{Labeled Faces in the Wild} (LFW) \cite{LFWTech} was chosen for its comprehensive collection of over 13\,000 images representative of everyday faces often exploited for identity-theft purposes. Its “in-the-wild” photography mirrors how source photos are scraped for zero-shot deepfake generators. However, as noted by its creators, LFW lacks demographic diversity, underrepresenting several ethnicities, age groups, and especially women.

To address the concern of fairness in the evaluation, this study utilizes \textbf{FairFace} \cite{karkkainenfairface}, a key dataset purposely curated for demographic diversity in facial analysis.
It is distinguished by its balanced representation of ethnicities, and age groups, encompassing a wide spectrum of human demographics.
Using FairFace and it's collection of attributes we can assess how each detector performs on a separate ethnic, age, and gender subset.

\subsection{Synthetic Datasets}
\label{sec:synth-data}

\begin{figure*}[tb]
  \centering
  \begin{subfigure}[b]{0.4\linewidth}
    \includegraphics[width=\linewidth]{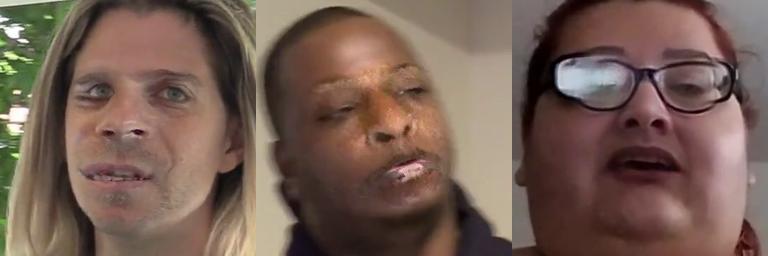}
    \caption{Samples from DFDC \cite{dolhansky2020deepfake}.}
  \end{subfigure}
  \begin{subfigure}[b]{0.4\linewidth}
    \includegraphics[width=\linewidth]{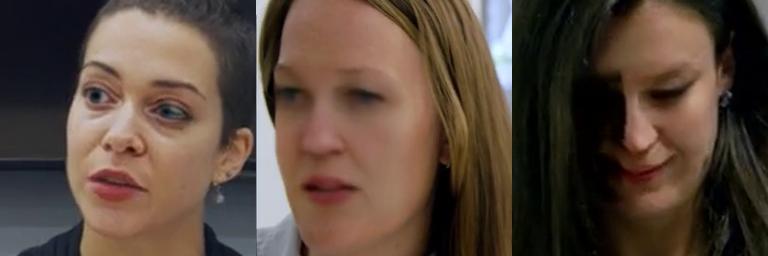}
    \caption{Samples from FaceForentics \cite{rossler2019faceforensics++}.}
  \end{subfigure}
    \caption{Many video frames from DFDC and FaceForentics datasets frequently demonstrate lower fidelity and a higher amount of artifacts due to specific keypoint matching and blending procedures as well as the usage of older deepfake methods.}
    \label{fig:bad_datasets}
\end{figure*}

Deepfake detectors often use prominent datasets such as Deepfake Detection Challenge (DFDC) \cite{dolhansky2020deepfake}, FaceForensics \cite{rossler2019faceforensics++}, and Celeb-DF \cite{li2020celeb}. However, the quality of the images in these datasets is questionable due to outdated image generation methods, low resolution, and other issues, as pointed out by the authors of Celeb-DF \cite{li2020celeb}. The DFDC and FaceForentics datasets largely contain deepfakes generated with pre-2019 models and do not reflect recent advances in deepfake technology. This raises concerns about the effectiveness of training and testing detectors with such data, potentially resulting in detectors that are ill-equipped to detect newer, more sophisticated deepfakes. In Figure \ref{fig:bad_datasets} we show some of the visual artifacts in the DFDC and FaceForentics datasets.

In response, we advocate modernizing test datasets with the latest deepfake generators. We use SimSwap \cite{10.1145/3394171.3413630} and Inswapper \cite{InsightFace} (also known as roop \cite{roop}), which power various face-swapping applications. We generate a new synthetic dataset by shuffling the target datasets and sampling two sets of images for face swapping, ensuring gender and, in the case of FairFace, age and race matching. This methodology results in a realistic and representative synthetic dataset, as shown in Figure \ref{fig:synth_data}.

For deepfake generation, we use standard preprocessing techniques with both SimSwap and Inswapper - cropping faces using the default bounding box selector and aligning faces using the default keypoint estimator. This simplicity of the generation process is a strength that allows for scalability and customization. Our approach can be extended to include more sophisticated deepfake models, additional labels and attributes. Using this methodology, we aim to bridge the gap in current deepfake detection capabilities by providing a robust, up-to-date testing framework that reflects advances in deepfake generation technologies. This not only improves the accuracy and reliability of the detector, but also encourages the research community to evolve and adapt test protocols as technology advances.

\subsection{Augmentations}
\label{sec:augmentations}

We propose a novel method to test deepfake detectors against real-world attacks by implementing augmentations that mimic the techniques used by fraudsters. Our approach divides these manipulative processes into two categories: artificial degradation and artificial enhancement.

\subsubsection{Artificial deterioration}

To mimic how attackers degrade image quality, we focus on structure-modifying augmentations while maintaining visual fidelity, excluding common training modifications such as blurring or color shifts. We employ two techniques: JPEG compression and downscaling.

\begin{figure*}[tb]
  \centering
  \begin{subfigure}[b]{0.3\linewidth}
    \includegraphics[width=\linewidth]{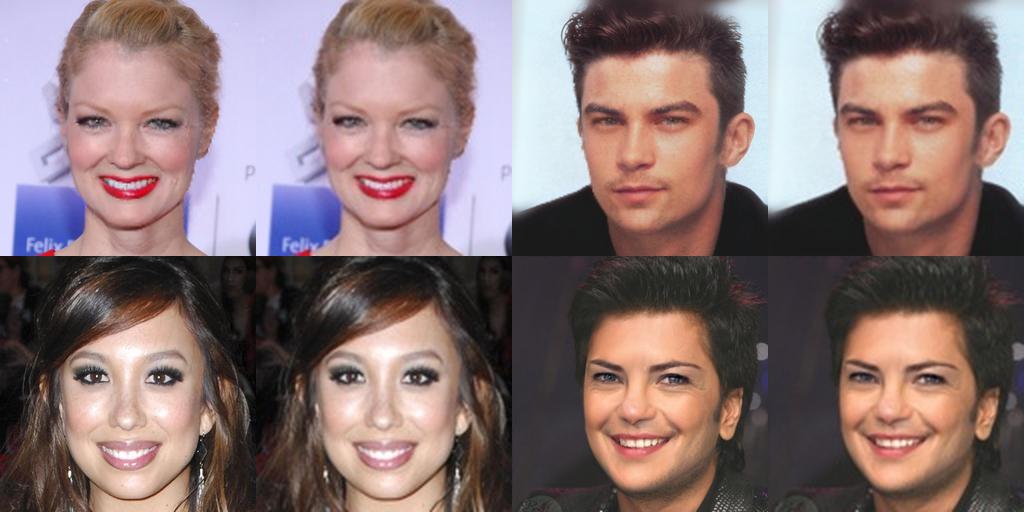}
    \caption{CelebA-HQ}
  \end{subfigure}
  \begin{subfigure}[b]{0.3\linewidth}
    \includegraphics[width=\linewidth]{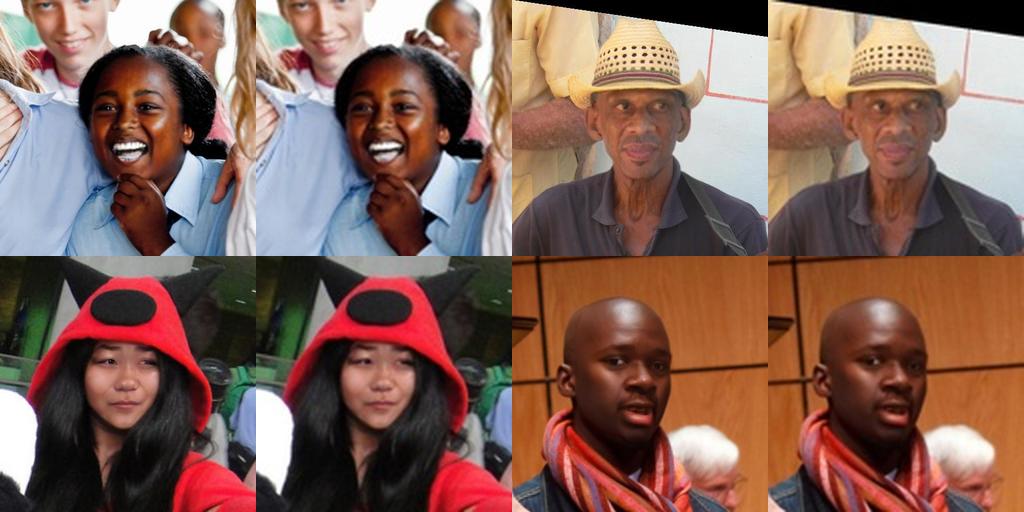}
    \caption{FairFace}
  \end{subfigure}
  \begin{subfigure}[b]{0.3\linewidth}
    \includegraphics[width=\linewidth]{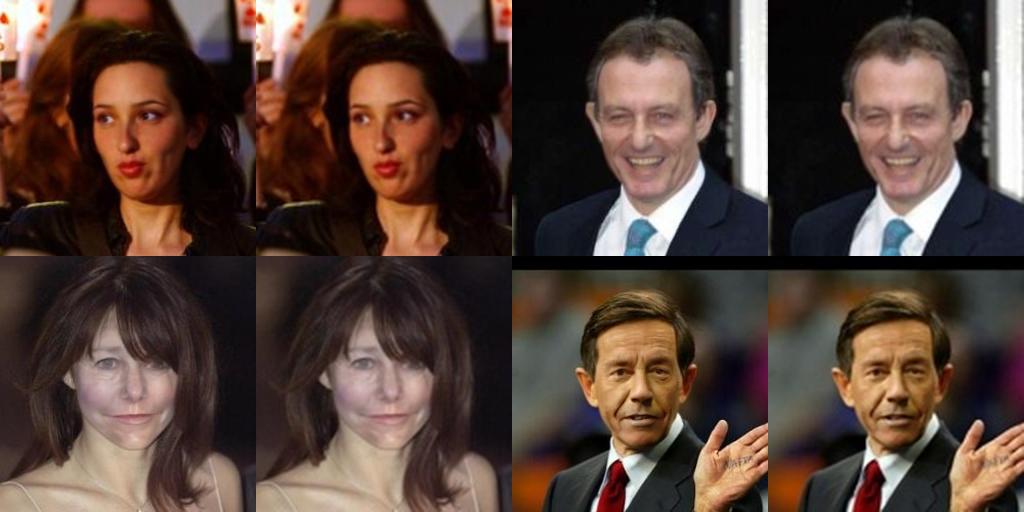}
    \caption{Labeled Faces in Wild}
  \end{subfigure}
    \caption{Examples of images from our synthetic dataset downscaled to 128 with the same aspect ratio.  Originals on the left; downscaled versions on the right. Notably, Inswapper (top row) samples are not heavily changed, while the deterioration of the SimSwap (bottom row) samples is visually noticeable.
    This degradation is attributed to SimSwap's native resolution of 512 pixels, which is higher than the downscaled size.}
    \label{fig:128_data}
\end{figure*}

\textbf{Downscale resizing} is used to simulate the effect of low-resolution recordings often seen in deepfake political attacks and attempts to bypass authentication systems. Most deepfake detectors include a resizing routine in their pre-processing, typically involving a face detector followed by cropping and resizing to a fixed size. While this preprocessing is standard, it can inadvertently degrade the detector's ability to distinguish between naturally low quality images and aggressively resized ones. This aspect is particularly important in know-your-customer (KYC) and liveness detection scenarios, where image quality can vary widely due to factors such as poor Internet connectivity or camera quality. Downscaled examples in our dataset, shown in Figure \ref{fig:128_data}, help to evaluate the performance of the detectors under these conditions.

\begin{figure*}[tb]
  \centering
  \begin{subfigure}[b]{0.4\linewidth}
    \includegraphics[width=\linewidth]{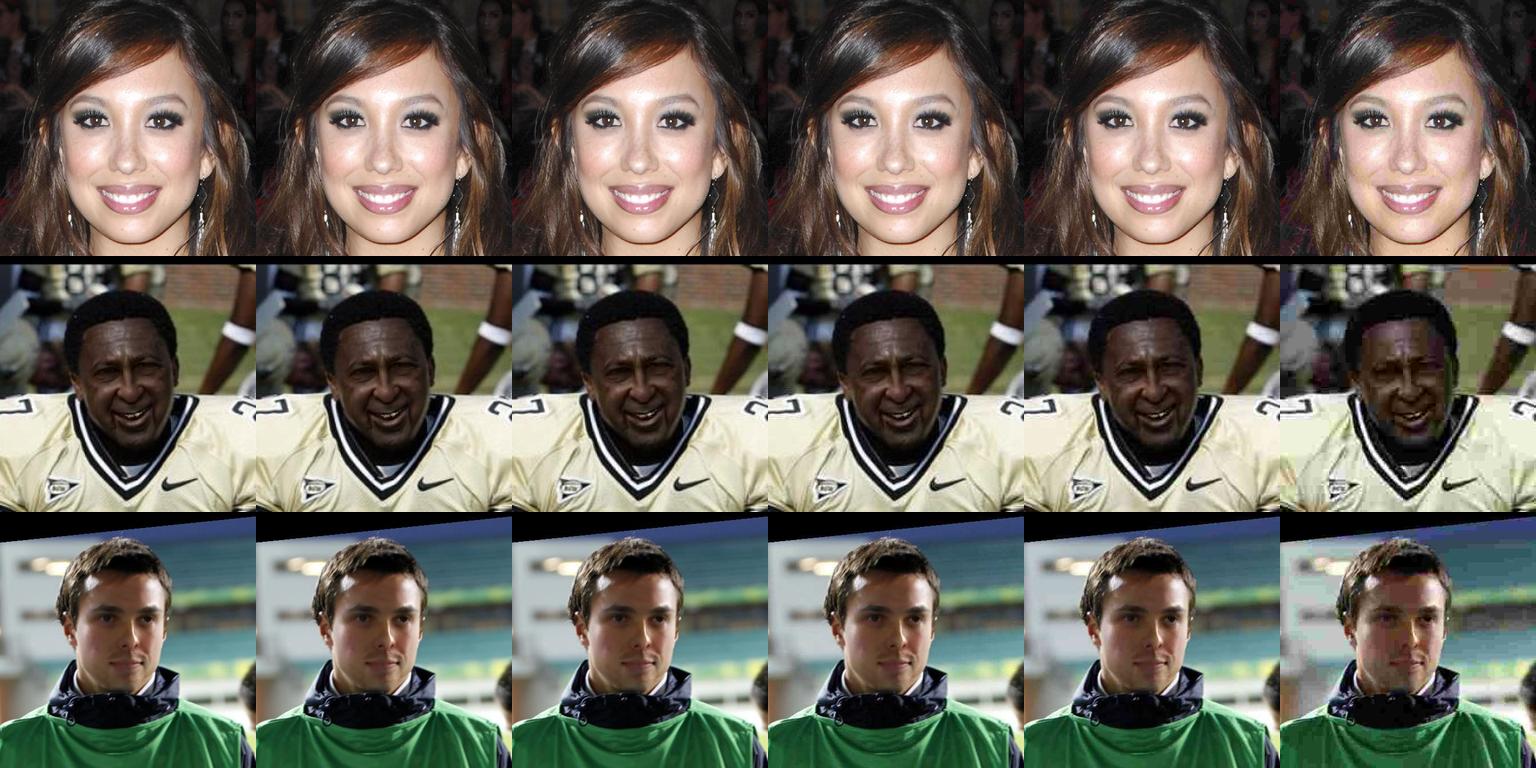}
    \caption{SimSwap}
  \end{subfigure}
  \begin{subfigure}[b]{0.4\linewidth}
    \includegraphics[width=\linewidth]{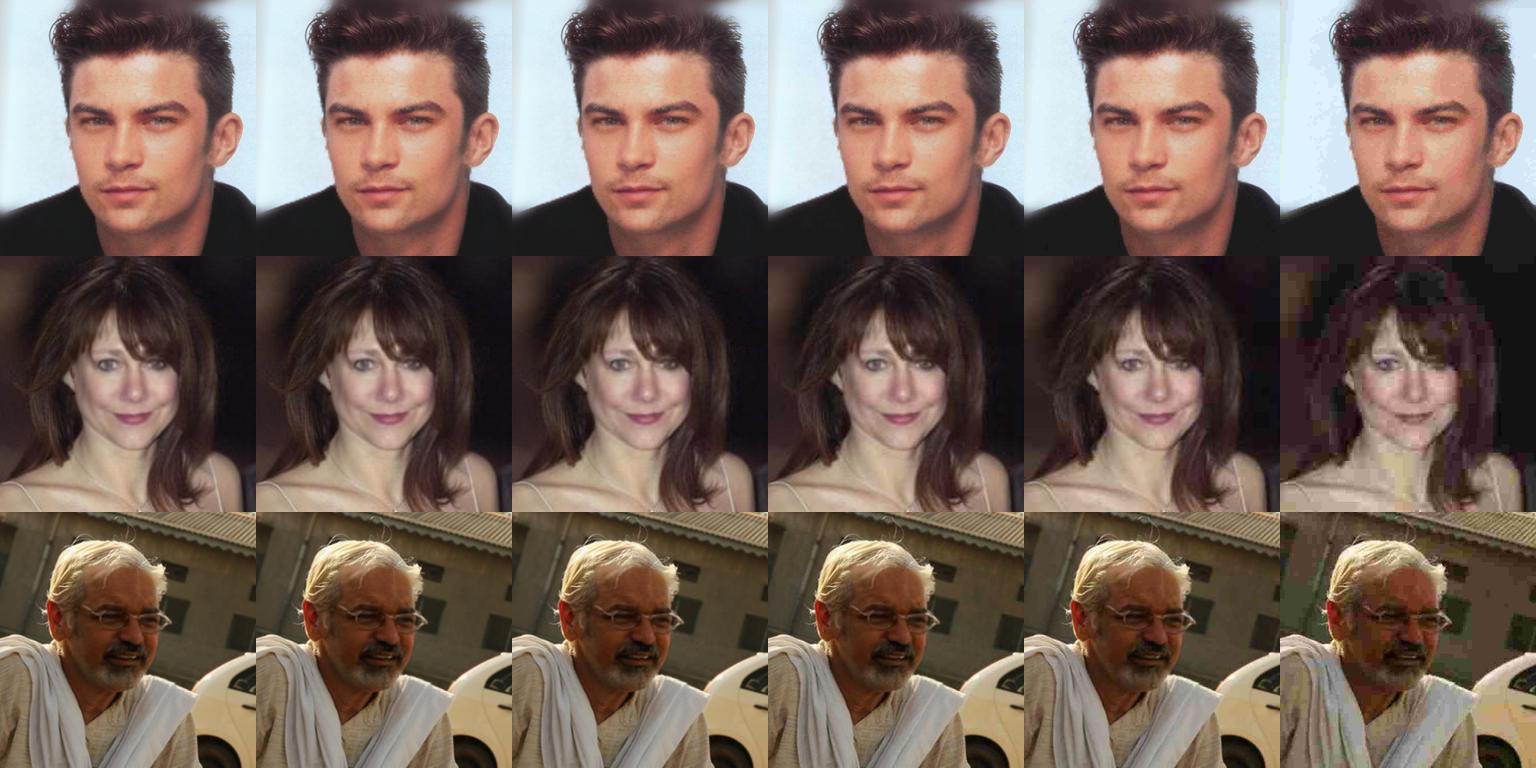}
    \caption{Inswapper}
  \end{subfigure}
  \caption{Visual examples of image quality deterioration for different JPEG compression coefficients. Coefficients from left to right: [original, 95, 75, 50, 30, 10]}
  \label{fig:jpeg_data}
\end{figure*}

\textbf{JPEG compression} tests the resilience of the detectors to changes in pixel values due to lossy compression, a common scenario in practical applications where uncompressed images are rarely used. This compression alters pixel values to reduce file size, resulting in minor detail loss and compression artifacts, as shown in Figure \ref{fig:jpeg_data}. A detector finely tuned to uncompressed image pixels may underperform on compressed images. Robustness to JPEG compression is a critical attribute that indicates the detector's ability to perform consistently regardless of the level of compression applied. This property is essential to ensure the detector's effectiveness in real-world conditions where image compression is commonplace.

\subsubsection{Artificial Enhancement.}

With the advancement of image enhancement tools such as GPEN \cite{yang2021gan}, fraudsters are increasingly refining deepfakes by enhancing facial features and textures. Therefore, it's crucial for deepfake detectors to be robust enough to handle both original generation methods and post-processing enhancements. To test this, we incorporate an artificial enhancement experiment into our testing pipeline, using the GPEN model to enhance images from our datasets. Figure \ref{fig:gpen_data} shows examples of GPEN-enhanced images, which we use to compare the robustness of the detector against enhancements. As with downscale resizing and JPEG compression, an effective detector should consistently detect deepfakes regardless of the artificial enhancements applied.

\section{Experiments}
\label{sec:experiments}

\begin{table*}[t]
\vskip 0.15in
\begin{center}
\begin{small}
\begin{sc}
\resizebox{\textwidth}{!}{%
  \begin{tabular}{|c|c|c|c|c|c|c|c|c|c|c|c|}
    \hline
    \multicolumn{2}{|c|}{} & \multicolumn{5}{c|}{SimSwap} & \multicolumn{5}{c|}{Inswapper} \\
    \hline
    Model & Dataset & ROC-AUC & F1  & PR-AUC & LogLoss & Accuracy & ROC-AUC  & F1  & PR-AUC  & LogLoss & Accuracy  \\
    \hline
    \hline
    \multirow{3}{*}{FF} & LFW          & 47.6 & 62.0 & 48.1 & 1.90 & 48.4 & 49.2 & 62.7 & 49.8 & 1.88 & 49.1 \\
                        & CelebA-HQ    & 58.9 & 53.3 & 62.7 & 1.00 & 57.5 & 78.5 & \textcolor{orange}{71.8} & \textcolor{orange}{79.7} & \textcolor{orange}{0.61} & \textcolor{orange}{70.7} \\
                        & FairFace     & 50.0 & 60.4 & 49.8 & 1.59 & 49.3 & 51.5 & 61.8 & 51.0 & 1.55 & 50.7 \\
                        & Overall      & 51.7 & 59.4 & 50.1 & 1.49 & 50.1 & 56.7 & 63.7 & 54.2 & 1.38 & 54.8 \\
    \hline
    \multirow{3}{*}{MAT} & LFW         & \textcolor{blue}{98.8} & \textcolor{blue}{88.2} & \textcolor{blue}{98.9} & 0.68 & \textcolor{blue}{86.8} & \textcolor{blue}{98.8} & \textcolor{blue}{88.2} & \textcolor{blue}{98.9} & \textcolor{blue}{0.68} & \textcolor{blue}{86.8} \\
                         & CelebA-HQ   & 49.0 & 66.6 & 48.9 & 0.69 & 50.0 & 49.8 & 66.6 & 49.6 & 0.69 & 50.0 \\
                         & FairFace    & 90.1 & 66.8 & 89.0 & \textcolor{green}{0.69} & 50.3 & \textcolor{green}{90.8} & 66.8 & \textcolor{green}{89.9} & \textcolor{green}{0.69} & 50.3 \\
                         & Overall     & 79.7 & 68.3 & 81.0 & \textcolor{yellow}{0.69} & 53.7 & \textcolor{yellow}{80.3} & 68.3 & \textcolor{yellow}{81.8} & \textcolor{yellow}{0.68} & 53.7 \\
    \hline
    \multirow{3}{*}{M2TR} & LFW        & 58.2 & 60.8 & 56.9 & 1.15 & 55.8 & 53.9 & 57.0 & 53.1 & 1.22 & 52.8 \\
                          & CelebA-HQ  & 56.3 & 52.5 & 55.1 & 1.18 & 54.4 & 57.1 & 53.6 & 55.9 & 1.16 & 55.1 \\
                          & FairFace   & 54.9 & 62.4 & 53.9 & 1.31 & 52.7 & 53.1 & 61.3 & 52.3 & 1.32 & 51.8 \\
                          & Overall    & 55.3 & 60.5 & 54.2 & 1.27 & 53.4 & 53.9 & 59.6 & 52.9 & 1.28 & 52.6 \\
    \hline
    \multirow{3}{*}{RECCE} & LFW       & 61.7 & 53.0 & 61.3 & 1.14 & 57.8 & 53.0 & 42.1 & 51.7 & 1.33 & 51.6 \\
                           & CelebA-HQ & 46.9 & 54.2 & 45.6 & 1.84 & 48.2 & 51.5 & 58.4 & 47.9 & 1.73 & 51.6 \\
                           & FairFace  & 60.1 & 64.1 & 59.0 & 1.53 & 54.6 & 58.5 & 64.0 & 56.1 & 1.53 & 54.6 \\
                           & Overall   & 56.7 & 61.3 & 54.2 & 1.56 & 53.5 & 56.1 & 61.5 & 53.1 & 1.55 & 53.6 \\
    \hline
    \multirow{3}{*}{CADDM} & LFW       & 87.8 & 76.8 & 87.9 & 0.86 & 71.4 & 58.2 & 58.7 & 59.2 & 1.19 & 55.6 \\
                           & CelebA-HQ & 75.2 & 68.6 & 74.6 & 0.95 & 68.7 & \textcolor{orange}{70.4} & 62.0 & 68.2 & 1.05 & 64.0 \\
                           & FairFace  & 79.2 & 73.6 & 77.4 & 0.98 & \textcolor{green}{68.0} & 56.6 & 58.3 & 55.7 & 1.24 & 54.9 \\
                           & Overall   & 78.2 & 73.0 & 77.6 & 0.96 & 68.5 & 59.8 & 59.1 & 58.5 & 1.20 & 56.9 \\
    \hline
    \multirow{3}{*}{SBI} & LFW         & 94.2 & 78.5 & 95.2 & \textcolor{blue}{0.64} & 73.9 & 78.3 & 72.5 & 78.3 & 0.75 & 67.8 \\
                         & CelebA-HQ   & \textcolor{orange}{93.6} & \textcolor{orange}{81.4} & \textcolor{orange}{93.4} & \textcolor{orange}{0.65} & \textcolor{orange}{77.7} & 70.0 & 64.1 & 64.1 & 0.93 & 65.5 \\
                         & FairFace    & \textcolor{green}{96.6} & \textcolor{green}{75.0} & \textcolor{green}{96.8} & 0.86 & 66.8 & 78.9 & \textcolor{green}{76.2} & 78.4 & 0.94 & \textcolor{green}{63.4} \\
                         & Overall   & \textcolor{yellow}{95.5} & \textcolor{yellow}{76.6} & \textcolor{yellow}{95.8} & 0.79 & \textcolor{yellow}{69.7} & 75.9 & \textcolor{yellow}{71.1} & 75.3 & 0.92 & \textcolor{yellow}{64.2} \\
    \hline
  \end{tabular}}
  \end{sc}
\end{small}
\end{center}
\vskip -0.1in
  \caption{
  The comparison in performance of deepfake detection models against SimSwap and Inswapper deepfake generators across three datasets: LFW, CelebA-HQ, and FairFace.
  Key performance indicators such as ROC-AUC, F1 score (with threshold of 0.5), Precision-Recall AUC, Log Loss, and Accuracy (with threshold of 0.5) are reported. 
  The highlighted numbers represent superior performance metrics, indicating an ability to classify deepfake samples from corresponding models.
}
  \label{table:compare_det}
\end{table*}

\begin{table*}[t]
\vskip 0.15in
\begin{center}
\begin{small}
\begin{sc}
\resizebox{\textwidth}{!}{%
  \begin{tabular}{|c|c|c|c|c|c|c|c|c|c|c|c|}
    \hline
    \multicolumn{2}{|c|}{} & \multicolumn{4}{c|}{SimSwap} & \multicolumn{4}{c|}{Inswapper} \\
    \hline
    Model & Dataset &
    Original & Low-quality & JPEG (75) & Enhanced & Original & Low-quality & JPEG (75) & Enhanced \\
    \hline
    \hline
    \multirow{3}{*}{FF}
    & LFW & 47.6 & 49.6 & 50.8 & 7.4 & 49.2 & 54.8 & 52.8 & 6.4 \\
    & CelebA & 58.9 & 78.1 & 61.0 & 26.2 & \textcolor{orange}{78.5} & 81.5 & 79.5 & 30.0 \\
    & FairFace & 50.0 & 60.0 & 53.7 & 18.6 & 51.5 & 59.4 & 53.8 & 18.8 \\
    & Overall & 51.7 & 62.9 & 54.9 & 20.2 & 56.7 & 63.0 & 58.7 & 20.8 \\
    \hline
    \multirow{3}{*}{MAT}
    & LFW & \textcolor{blue}{98.8} & 35.4 & \textcolor{blue}{100.0} & 35.6 & \textcolor{blue}{98.9} & 32.9 & \textcolor{blue}{100.0} & \textcolor{blue}{100.0} \\
    & CelebA & 49.0 & \textcolor{orange}{89.6} & 72.1 & 82.5 & 49.8 & \textcolor{orange}{88.9} & \textcolor{orange}{83.3} & 78.5 \\
    & FairFace & 90.1 & \textcolor{green}{90.6} & \textcolor{green}{90.4} & \textcolor{green}{82.2} & \textcolor{green}{90.8} & \textcolor{green}{90.1} & \textcolor{green}{90.5} & \textcolor{green}{80.6} \\
    & Overall & 79.7 & \textcolor{yellow}{83.0} & \textcolor{yellow}{88.1} & 76.3 & \textcolor{yellow}{80.3} & \textcolor{yellow}{82.6} & \textcolor{yellow}{90.2} & \textcolor{yellow}{82.2} \\
    \hline
    \multirow{3}{*}{M2TR}
    & LFW  & 58.2 & \textcolor{blue}{65.8} & 58.3 & 47.7 & 53.9 & 59.8 & 53.9 & 41.0 \\
    & CelebA & 56.3 & 64.3 & 56.3 & 75.3 & 57.1 & 65.2 & 57.1 & 75.1 \\
    & FairFace & 54.9 & 63.1 & 55.0 & 46.9 & 53.1 & 60.0 & 53.1 & 43.8 \\
    & Overall & 55.3 & 63.1 & 55.3 & 53.3 & 53.9 & 60.8 & 53.9 & 50.6 \\
    \hline
    \multirow{3}{*}{RECCE}
    & LFW  & 61.7 & 62.0 & 58.7 & \textcolor{blue}{82.3} & 53.0 & \textcolor{blue}{60.0} & 64.5 & 77.5 \\
    & CelebA & 46.9 & 28.2 & 52.1 & 80.3 & 51.5 & 33.6 & 63.3 & 82.0 \\
    & FairFace & 60.1 & 38.0 & 51.5 & 63.8 & 58.5 & 37.1 & 58.6 & 62.1 \\
    & Overall & 56.7 & 38.2 & 52.2 & 69.4 & 56.1 & 38.7 & 60.1 & 68.1 \\
    \hline
    \multirow{3}{*}{CADDM}
    & LFW  & 87.8 & 63.0 & 70.5 & 73.1 & 58.2 & 50.0 & 50.3 & 69.2 \\
    & CelebA & 75.2 & 74.4 & 63.7 & 56.8 & 70.4 & 75.4 & 67.1 & 57.1 \\
    & FairFace & 79.2 & 59.0 & 69.8 & 59.6 & 56.6 & 49.8 & 55.1 & 51.4 \\
    & Overall & 78.2 & 62.7 & 67.9 & 59.8 & 59.8 & 56.0 & 57.3 & 54.1 \\
    \hline
    \multirow{3}{*}{SBI}
    & LFW  & 94.2 & 43.2 & 73.8 & 71.8 & 78.3 & 33.4 & 51.6 & 68.4 \\
    & CelebA & \textcolor{orange}{93.6} & 41.1 & \textcolor{orange}{90.6} & \textcolor{orange}{86.7} & 70.0 & 45.3 & 68.1 & \textcolor{orange}{82.9} \\
    & FairFace & \textcolor{green}{96.6} & 37.7 & 85.6 & 75.2 & 78.9 & 29.7 & 64.8 & 69.1 \\
    & Overall & \textcolor{yellow}{95.5} & 38.6 & 85.5 & \textcolor{yellow}{77.7} & 75.9 & 33.5 & 63.9 & 72.4 \\
    \hline
  \end{tabular}}
  \end{sc}
\end{small}
\end{center}
\vskip -0.1in
  \caption{\em \small AUC-ROC comparison on performance of selected detectors when evaluating SimSwap and Inswapper generated images across different quality modifications.
  The models were tested on their ability to distinguish between original, low-quality, JPEG compressed, and GPEN-enhanced images.
}
  \label{table:compare_augs}
\end{table*}

In this section, we showcase a set of experiments conducted to assess the effectiveness of the six chosen deepfake detection models described in Section \ref{sec:detectors}.
Each model was created using code from an official GitHub repository with a corresponding set of weights and hyperparameters according to the instructions provided by the authors. 
Each experiment and each model were run in a specific environment with fixed Python libraries and CUDA versions that matched the ones disclosed by the authors.

For our experiments, we utilized three widely recognized datasets: LFW, CelebA-HQ, and FairFace, with reasons for their selection detailed in Section \ref{sec:datasets}. We ensured that the six selected models were not trained on any of these datasets, supporting the generalizability of our evaluation pipeline. Even if a model was trained on CelebA-HQ or LFW, our pipeline's robustness against attacks demonstrates its in-dataset stability. To evaluate out-of-domain performance, we incorporated the FairFace dataset, which has more specific applications and is less common in deepfake detector training. This approach, including the potential to add any high-fidelity dataset, exemplifies the versatility of our proposed methodology.

\subsection{One Sided Testing}

\begin{figure*}[ht]
\vskip 0.2in
\begin{center}
\centering
\resizebox{\textwidth}{!}{%
\begin{subfigure}{.3\textwidth}
  \centering
  \includegraphics[width=\linewidth]{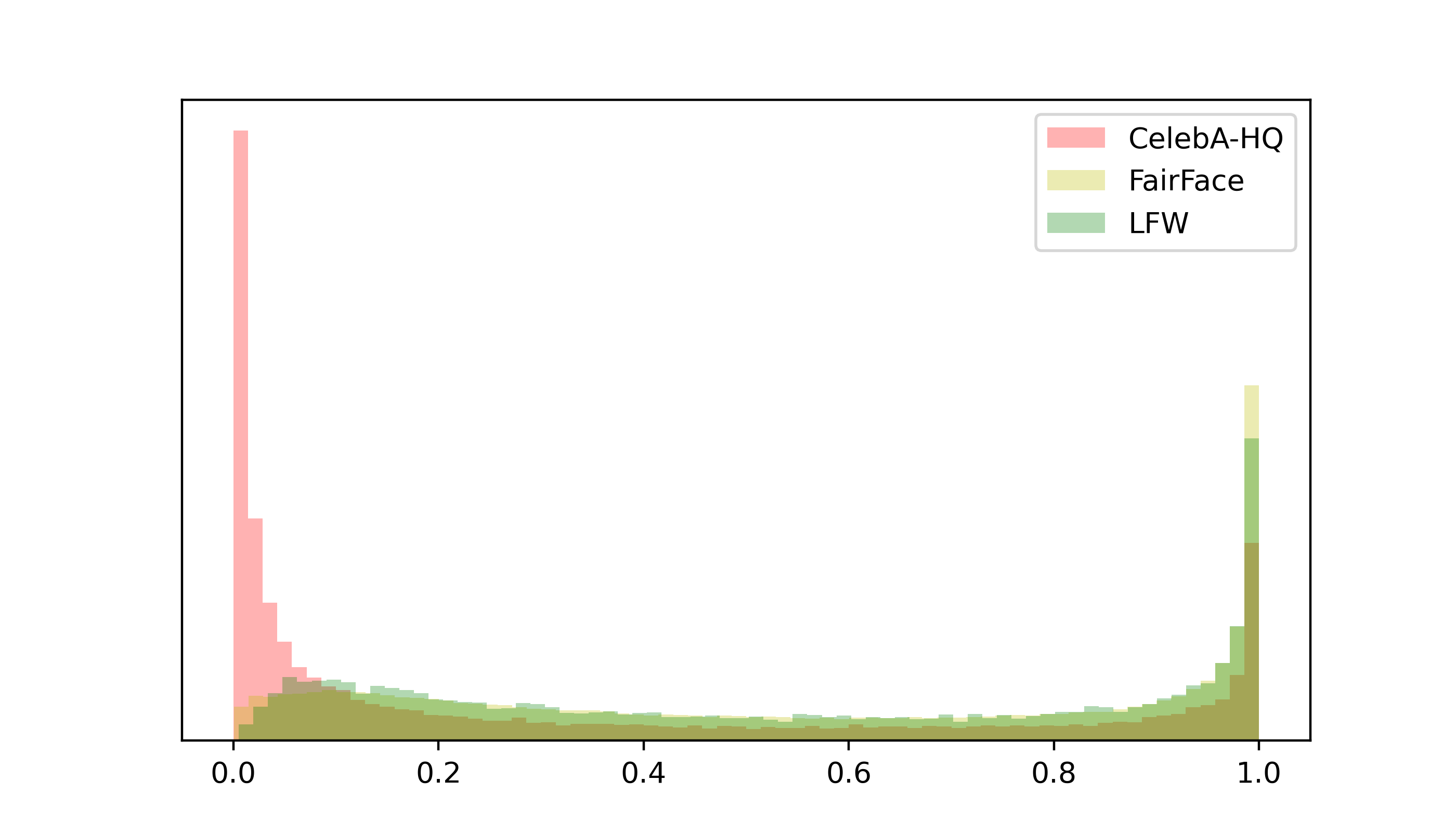}
  \caption{CADDM}
\end{subfigure}
\begin{subfigure}{.3\textwidth}
  \centering
  \includegraphics[width=\linewidth]{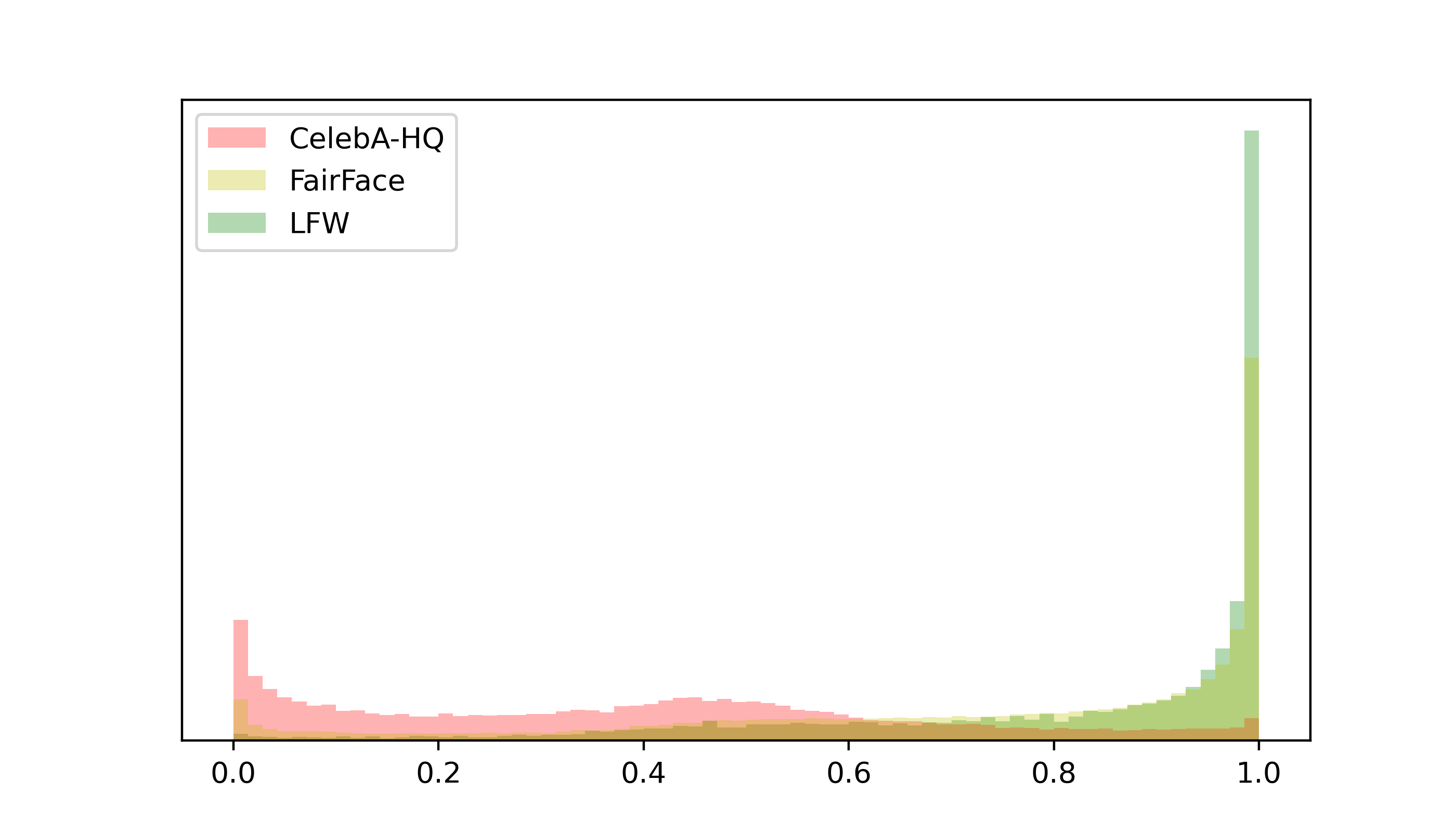}
  \caption{FF}
\end{subfigure}
\begin{subfigure}{.3\textwidth}
  \centering
  \includegraphics[width=\linewidth]{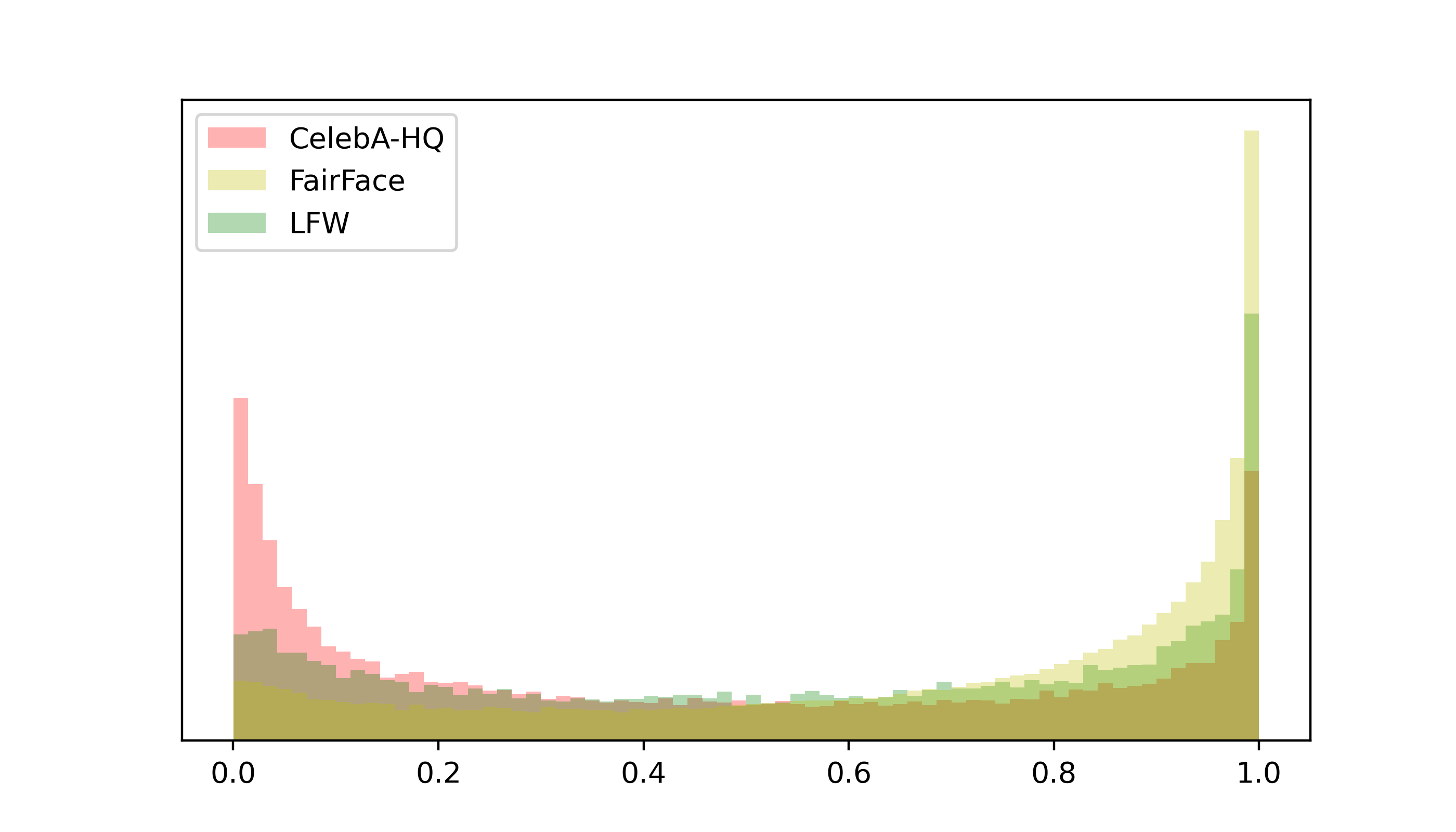}
  \caption{M2TR}
\end{subfigure}}

\resizebox{\textwidth}{!}{%
\begin{subfigure}{.3\textwidth}
  \centering
  \includegraphics[width=\linewidth]{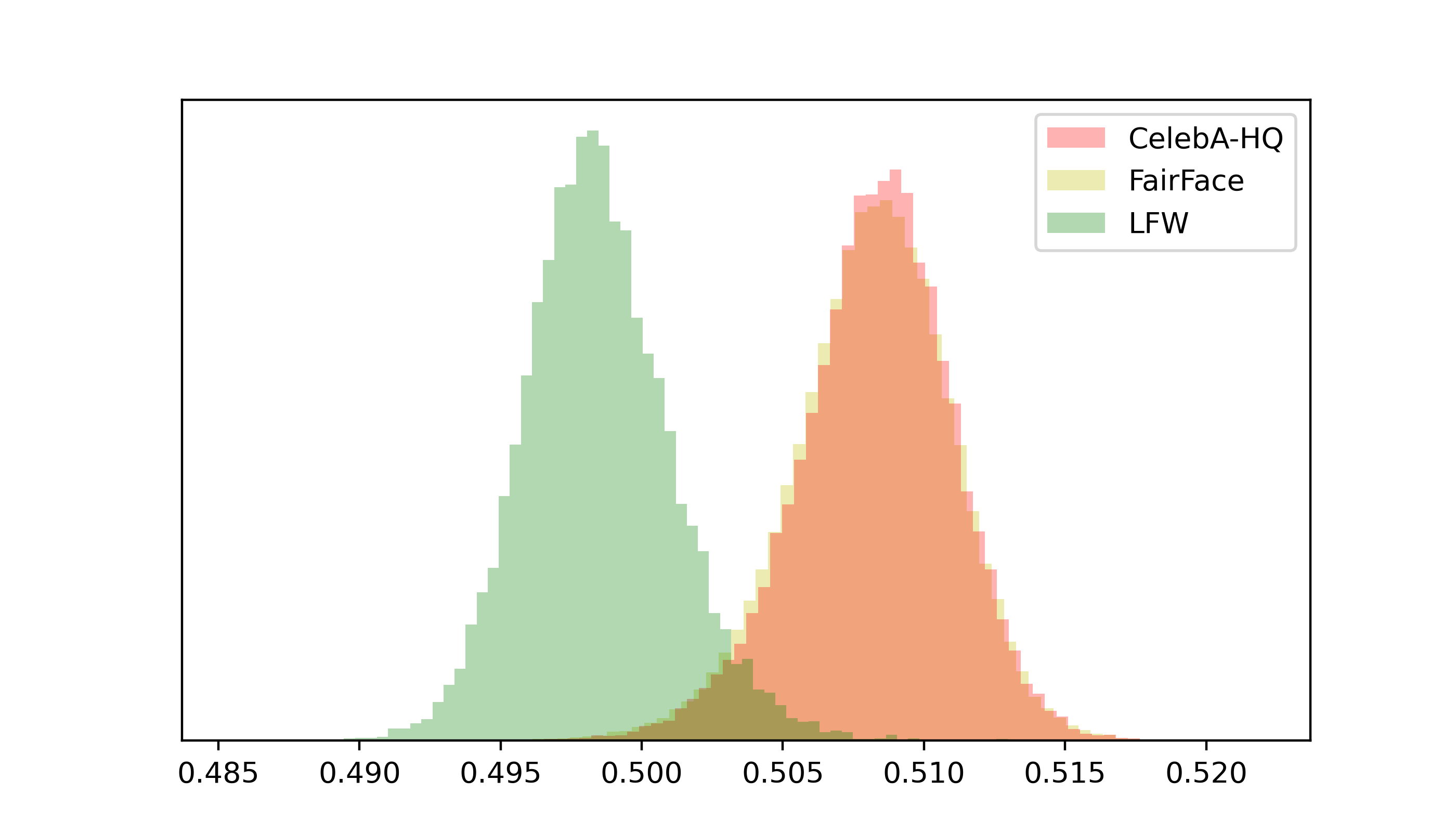}
  \caption{MAT}
\end{subfigure}
\begin{subfigure}{.3\textwidth}
  \centering
  \includegraphics[width=\linewidth]{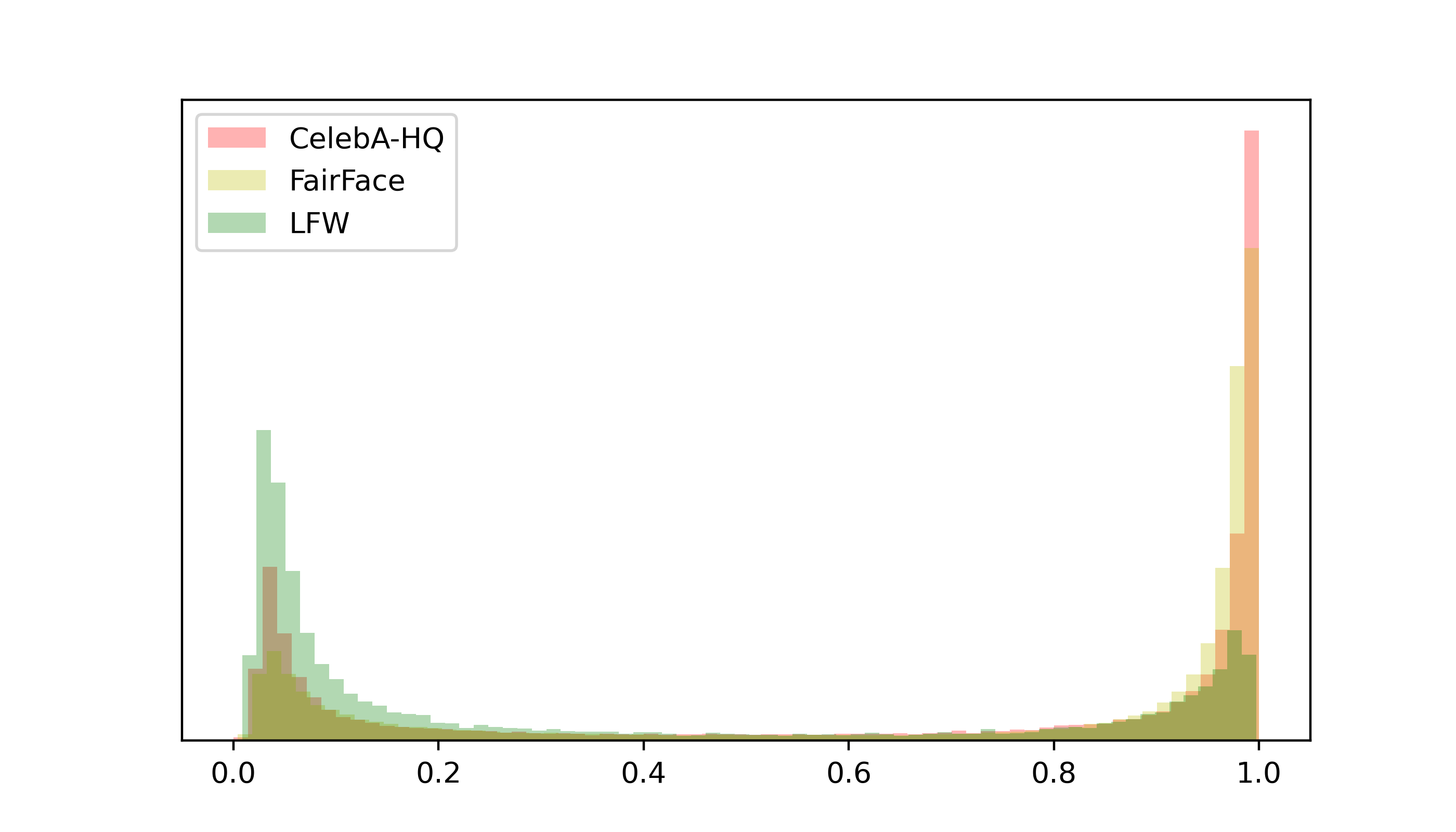}
  \caption{RECCE}
\end{subfigure}
\begin{subfigure}{.3\textwidth}
  \centering
  \includegraphics[width=\linewidth]{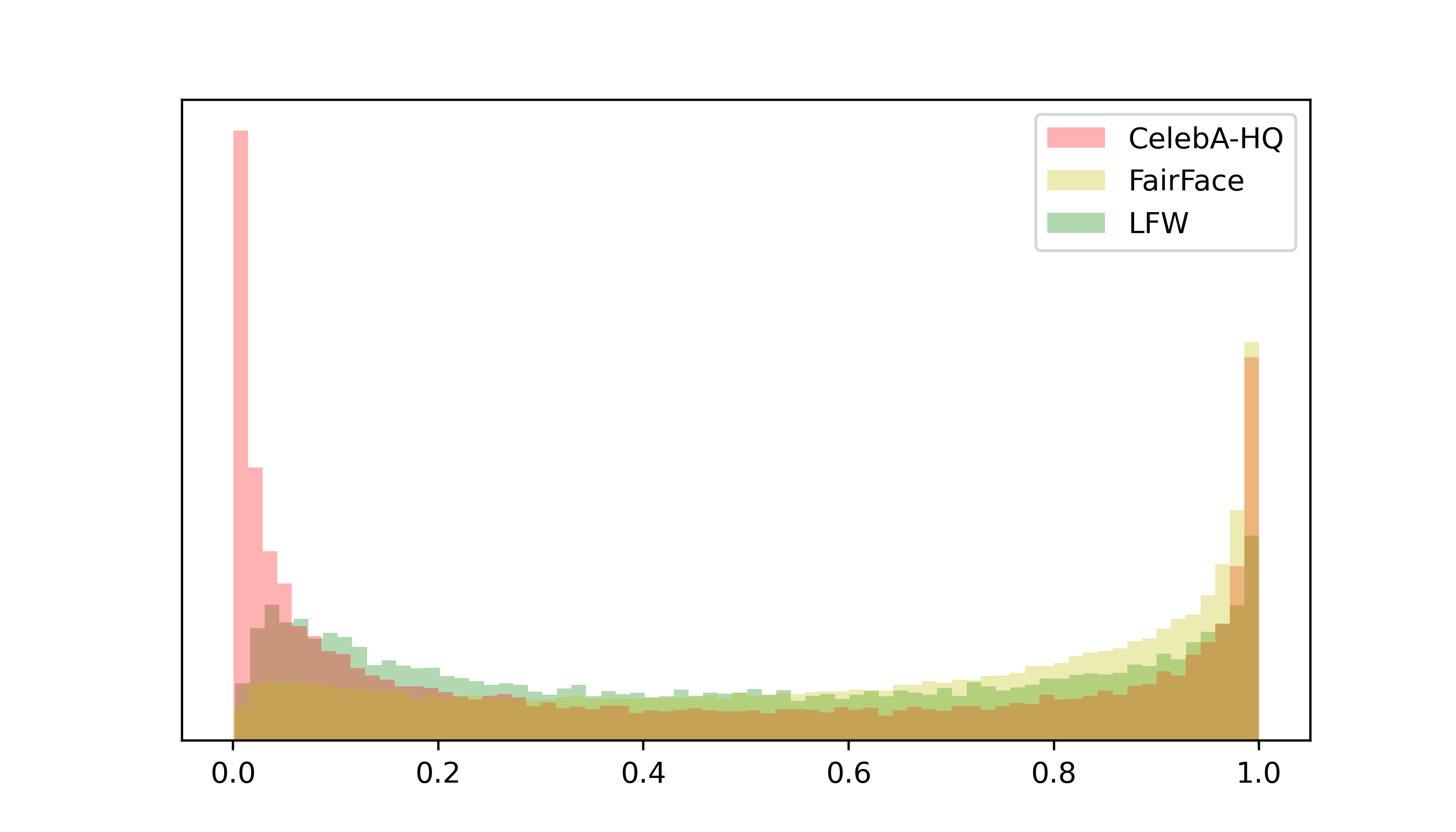}
  \caption{SBI}
\end{subfigure}}
\caption{Normalized histograms of predicted values for "real" images in selected datasets. 
Here one can see that all models failed to classify "real" samples with some incorrectly assigning high probability to a major percentage of dataset samples.}
\label{fig:hists_real}
\end{center}
\vskip -0.2in
\end{figure*}

This experiment evaluated the initial quality of different deepfake detectors using only real images from the CelebA-HQ, LFW, and FairFace datasets. We applied each model to these images and analyzed the distribution of predicted values, shown in Figure \ref{fig:hists_real}, where zero indicates "not deepfake" and one indicates "deepfake". Contrary to expectations, where predictions should cluster around zero, the results varied significantly between models. Some models performed well on one dataset, but poorly in others.
This finding raises concerns about the ability of models to generalize beyond their training domain. 
This test shows that most of the selected models frequently misclassified real images as deepfakes, assigning high probabilities close to one, indicating a significant gap in their practical applicability.

\subsection{Classification}

To properly assess the models' quality, we calculate classic classification metrics, such as ROC-AUC, PR-AUC, logloss, F1 score, and accuracy. 
Because each detector was trained differently and the scale of prediction values can vary, we advise against using the F1 score and accuracy with default 0.5 threshold as a main comparison metric. 
Instead, we look to ROC-AUC as the main metric, as it evaluates the performance of a classification model at all possible classification thresholds and provides a comprehensive measure of the model's ability to distinguish between the two classes.

We employ synthetic datasets described in \ref{sec:synth-data} and label all images from them as "fake".
The results can be seen in Table \ref{table:compare_det}.
It would be safe to say that from all selected models only MAT, SBI, and CADDM models are showing high enough quality in detecting deepfakes from SimSwap and Inswapper on all three datasets.
Each of the six models introduces innovative approaches within the field. 
However, we argue that only models that can effectively disentangle learned knowledge from a training dataset (such as SBI model) can achieve high-quality performance on unseen datasets, as evidenced by this experiment.

\subsection{JPEG compression attack}

\begin{figure*}[h!]
\vskip 0.2in
\begin{center}
\resizebox{\textwidth}{!}{%
\begin{subfigure}{.3\textwidth}
  \centering
  \includegraphics[width=\linewidth]{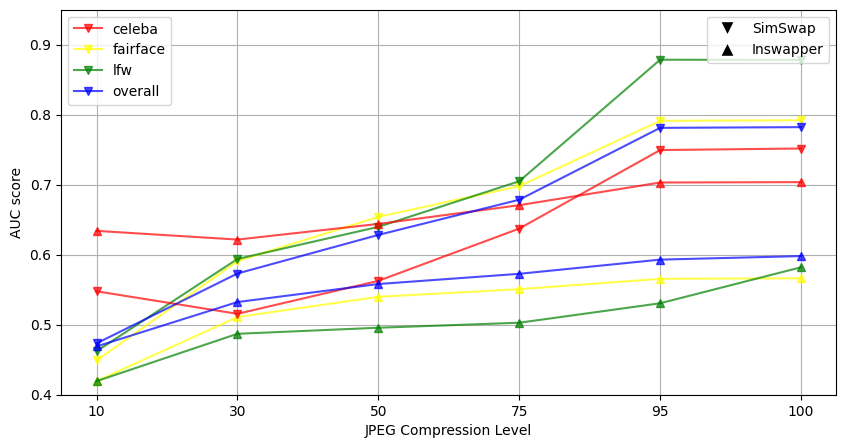}
  \caption{CADDM}
\end{subfigure}%
\begin{subfigure}{.3\textwidth}
  \centering
  \includegraphics[width=\linewidth]{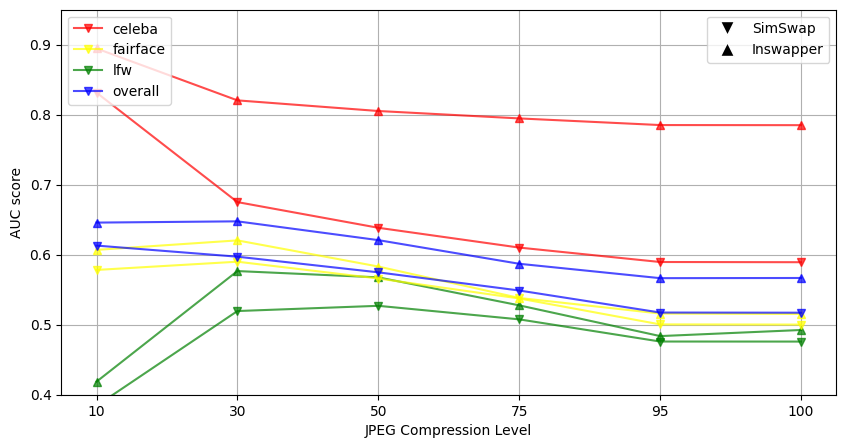}
  \caption{FF}
\end{subfigure}
\begin{subfigure}{.3\textwidth}
  \centering
  \includegraphics[width=\linewidth]{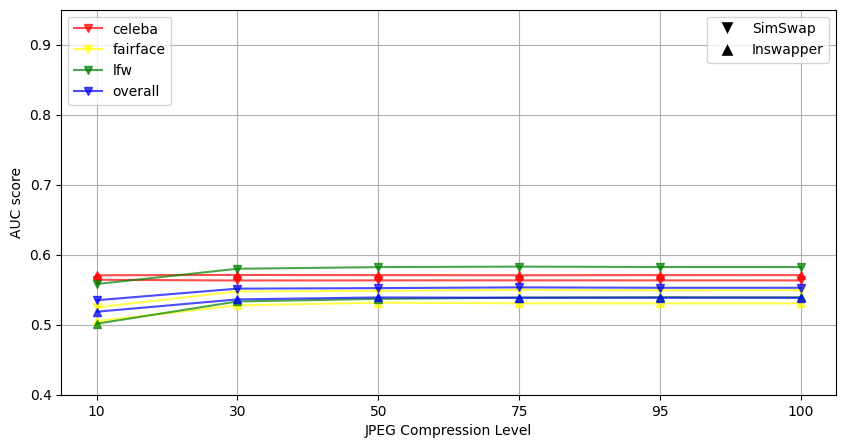}
  \caption{M2TR}
\end{subfigure}}

\resizebox{\textwidth}{!}{%
\begin{subfigure}{.3\textwidth}
  \centering
  \includegraphics[width=\linewidth]{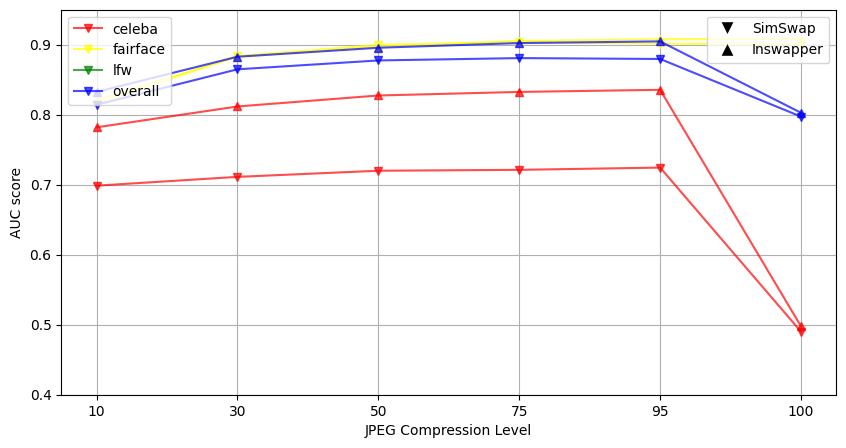}
  \caption{MAT}
\end{subfigure}
\begin{subfigure}{.3\textwidth}
  \centering
  \includegraphics[width=\linewidth]{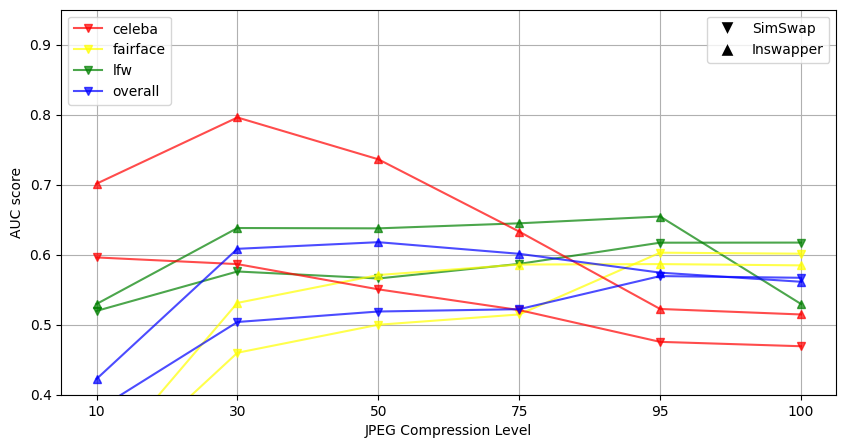}
  \caption{RECCE}
\end{subfigure}
\begin{subfigure}{.3\textwidth}
  \centering
  \includegraphics[width=\linewidth]{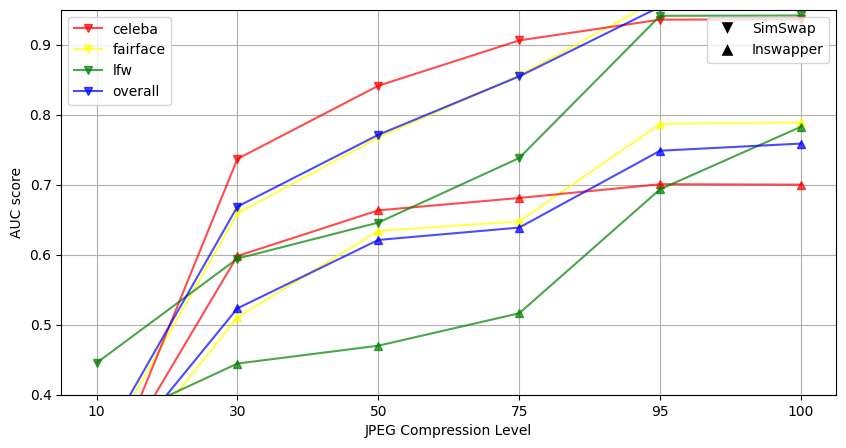}
  \caption{SBI}
\end{subfigure}}
\caption{AUC-ROC by JPEG compression coefficient on all models, separated by datasets.}
\label{fig:jpeg_comp}
\end{center}
\vskip -0.2in
\end{figure*}

The next step of the proposed testing pipeline consists of evaluating detectors against the simplest yet visually undetectable attack: JPEG compression.
The experimental setup mirrors the prior one, with the significant distinction of subjecting each image in the synthetic datasets to JPEG compression at various quality levels: 95, 75, 50, 30, and 10.
An ideal detector, similar to a human moderator, should be robust against this attack, since the visual context in the image stays the same before and after the attack.
However, in cases where the detector is relying too much on exact pixel values seen in the training dataset, it could become unstable after JPEG compression attack.
In Figure \ref{fig:jpeg_data} we show examples of images before and after such attacks with different compression coefficients. 
Figure \ref{fig:jpeg_comp} demonstrates how the ROC-AUC of selected detectors changes with different JPEG compression coefficients.
We find it interesting that for some detectors, such as SBI, the increasing compression rate decreases the ROC-AUC.
While for other detectors, like FF and MAT, it provides a signal necessary to score all compressed images as fakes, increasing the quality as the compression rate decreases.

\subsection{Low and High-Quality Attacks}

This section evaluates the response of modern deepfake detectors to evasion tactics, focusing on extremes of feed quality, as artificially lowering image quality can help deepfakes evade moderation. To test "low-quality attacks," we downscaled our synthetic dataset images to a maximum of 128 pixels while maintaining the aspect ratio, as shown in Figure \ref{fig:128_data}. We then evaluated the ability of the detectors to discriminate between original and downscaled deepfakes. Most models, especially RECCE and SBI, struggled with the downscaling, as detailed in Table \ref{table:compare_augs}. The SBI model, designed to detect blending artifacts, and RECCE, which focuses on forgery signatures, failed to detect these post-downscaled enhancements.

Conversely, high-quality enhancements are less suspicious, but equally challenging for detectors. We used the GPEN enhancer, a state-of-the-art model capable of real-time processing, to improve the quality of synthetic images. The enhanced images were then tested for detector vulnerability, with the results presented in Table \ref{table:compare_augs}. The enhanced images significantly changed the results, with some models' ROC-AUCs dropping below 50, indicating misclassification of deepfakes as real. Interestingly, models such as RECCE showed improved metrics after enhancement.

\section{Results}

In our experimental analysis, we focused on evaluating the performance of deepfake detection models against varying manipulations applied to deepfake images.
The results, as summarized in Table \ref{table:compare_det}, indicate that some state-of-the-art detectors are not robust enough to correctly classify samples from modern deepfake generators.
Models such as RECCE, FF, and M2TR while demonstrating great quality in their original papers, completely fail to distinguish real and fake samples in our pipeline.
At the same time, the SBI model performs greatly on all three datasets, followed by the CADDM and MAT models.

From Table \ref{table:compare_augs}, it was evident that all models demonstrated a certain level of vulnerability when processing augmented images, with RECCE and CADDM being the most affected by downscaled images.
Similarly, the SBI model, while showing high accuracy under original conditions, experienced a marked decline in performance after images were subjected to downscaling and enhancement, indicating a sensitivity to alterations in image quality.

Interestingly, the application of the GPEN enhancer resulted in an improved, yet varied, detection rate.
The performance of the RECCE and CADDM models demonstrate that there may be a relative increase in performance when analyzing enhanced images, demonstrating a partial skew in performance towards images of a higher resolution. 
Similar behavior was exhibited by MAT model on the CelebA-HQ dataset specifically.
Additionally, the performance of the SBI model remained significantly lower on enhanced images, highlighting its sensitivity to the quality of input data.

Overall, our findings illustrate the complexities deepfake detectors face when encountering synthetic images, especially ones that have been deliberately altered in quality.
From Tables \ref{table:compare_det} and \ref{table:compare_augs} it is clear that no single model consistently maintains high performance metrics such as ROC-AUC, across all variations of image manipulations and datasets.
One notable exception to this is the SBI model, but despite its higher-than-average ROC-AUCs there is a significant drop in performance when images are downscaled. 
At the same time, models like FF and M2TR demonstrate low success rates even on unchanged datasets, as evidenced by their lower scores in precision and log-loss metrics.

The instability of the selected deepfake detectors is further highlighted by their inconsistent performance across different datasets.
For instance, the MAT model scores vary greatly between the LFW and CelebA-HQ datasets, suggesting a lack of generalizability.
This suboptimal performance across various datasets can be largely attributed to overtraining on features specific to training datasets.
This trend is common throughout our findings, underscoring the current limitations of deepfake detectors.

\section{Conclusion}
In this work, we show that deepfake detection models, although robust in controlled tests, falter in real-world scenarios.
We developed a flexible testing pipeline that has highlighted these limitations, indicating a need for models that can withstand various adversarial strategies.
We want to emphasize that there is \textbf{no ultimate solution} for deepfake detection, and to aid in ongoing research, we have shared our codebase along with a substantial deepfake dataset on GitHub for community use.
This effort aims to stimulate advancements in detection technology and inform regulatory approaches.

\noindent\textbf{Limitations.} There is still room for improvement in the proposed pipeline. The main problem is the abundance of deepfake detection methods with unpublished code or weights, preventing comprehensive testing of their efficacy. Another issue is the lack of publicly available identity verification datasets, where deepfake detection remains a significant challenge. We look forward to addressing both of these gaps in future work.

\bibliography{main}
\bibliographystyle{icml2025}

\end{document}